\documentclass[11pt]{article}
\usepackage{geometry}
\usepackage{coling2020}
\usepackage{times}
\usepackage{url}
\usepackage{latexsym}
\usepackage{microtype}
\usepackage{graphicx}
\usepackage{amsmath}

\colingfinalcopy 

\title{NITS-Hinglish-SentiMix  at SemEval-2020 Task 9: Sentiment Analysis For Code-Mixed Social Media Text Using an Ensemble Model\\
}

 \author{Subhra Jyoti Baroi \hspace{12pt} Nivedita Singh \hspace{12pt} Ringki Das    \hspace{12pt} Thoudam Doren Singh\\
 Department of Computer Science and Engineering \\
 National Institute of Technology Silchar\\
 Silchar, Assam -788010\\
 {\tt \{tushar.barroi99,s.nivedita279,ringkidas,thoudam.doren\}@gmail.com}
 }
 
\date{}

\begin{document}
\maketitle
\begin{abstract}

Sentiment Analysis is the process of deciphering what a sentence emotes and classifying them as either positive, negative, or neutral. In recent times, India has seen a huge influx in the number of active social media users and this has led to a plethora of unstructured text data. Since the Indian population is generally fluent in both Hindi and English, they end up generating code-mixed Hinglish social media text i.e. the expressions of Hindi language, written in the Roman script alongside other English words. The ability to adequately comprehend the notions in these texts is truly necessary. Our team, \textbf{rns2020} participated in Task 9 at SemEval2020 intending to design a system to carry out the sentiment analysis of code-mixed social media text. This work proposes a system named NITS-Hinglish-SentiMix to viably complete the sentiment analysis of such code-mixed Hinglish text. The proposed framework has recorded an F-Score of 0.617 on the test data.   

\end{abstract}

\section{Introduction}

\blfootnote{
    
    \hspace{-0.65cm}  
    This work is licensed under a Creative Commons 
    Attribution 4.0 International Licence.
    Licence details:
    \url{http://creativecommons.org/licenses/by/4.0/}.
    
}

With the advent of social media, India alone stood at 326.1 million users in 2018. A large portion of the population is fluent in both English and Hindi henceforth it is regular to see the use of code-mixed Hinglish language on social media platforms. In the world of natural language processing(NLP), sentiment analysis of the code-mixed Hinglish text is a great challenge. The task at hand is the SemEval-2020 Task 9 \cite{patwa2020sentimix}: Sentiment Analysis for Code-Mixed Social Media Text for Hinglish. Code-mixed Hinglish tweets are provided as training, validation, and test sets for model training, validation, and testing respectively. The proposed framework NITS-Hinglish-SentiMix, an ensemble model wherein different models have been consolidated to improve the general F-Score of the framework. The referenced group model is a mix of a basic LSTM (Long Short Term Memory), an LSTM+Convolution, a BiLSTM (Bidirectional LSTM), and a CNN (Convolution Neural Network) model. This paper is composed as follows, Section 2 is a concise writing overview of related works. Section 3 talks about the strategy which covers bits of insights into the dataset given, alongside the pre-processing, the details on the model architecture. Section 4 talks about the training and results acquired and the paper concludes with Section 5.

\section{Literature Survey}

In recent years, a lot has changed in the field of sentiment analysis. This section discusses a few of the works which helped us build insight into the problem and come up with a new solution. The initial approach towards sentiment analysis includes systems powered by traditional machine learning techniques like the work by \cite{Agarwal+Bhattacharyya:05a} explores the performance of SVM for the task of multinomial classification of documents. The approach used the WordNet synonym graph with good and bad as anchor points to find out the mutual relationship of words between the documents and the sentences.These relationships were then used for sentiment prediction with the help of SVM. Graph cut techniques were also applied to improve classification accuracy. The work by \cite{wang-etal-2012-system} proposes a real-time system for sentiment analysis of tweets made of many steps but only 2 steps were of prime importance i.e pre-processing and sentiment prediction. Pre-processing includes tokenization with a tokenizer that handled the URLs, common emoticons, phone numbers, HTML tags, twitter mentions and hashtags, numbers with fractions and decimals, repetition of symbols and Unicode characters dexterously and ensured no noise in the processed data. The statistical classifier used for sentiment analysis or prediction was a Naïve Bayes model on unigram features. The architecture and the methods used were generic and hence the domain of this model could be extended. The paper proposed by \cite{mathur-etal-2018-detecting} focuses on the classification of offensive tweets written in the Hinglish language. They approach the problem of sentiment analysis or classification of the tweets in the HEOT (Hindi-English Offensive Tweet) dataset using the concepts of transfer learning where the proposed model consists of convolutional neural networks (CNN) pre-trained English tweets followed by retraining on Hinglish tweets. Their paper also examined various CNN based models for the task but leaves out other deep learning models based on LSTM and BiLSTM which are expected to show a high affinity towards semantic-based tasks like sentiment analysis. The works proposed by \cite{kenyon-dean-etal-2018-sentiment}  helped in manually annotating the data as it provides an insight into how humans annotate data. The study was conducted on a dataset specifically prepared for this purpose called the McGill Twitter Sentiment Analysis (MTSA) dataset containing 7,026 tweets. Translating code-mixed social media comments is attractive and challenging research. Work by \cite{singh2017towards} suggested a translation model on code-mixed Facebook comments. The proposed system was based on two approaches, one is language identified and without using language identifier. The Hindi-English code-mixed model has achieved an improved result over the baseline model. Code-switching is an alternation of spoken language i.e. utterance or conversation in the multilingual community. In various natural language processing tasks like named entity tasks, parts of speech tagging, sentiment analysis, machine translation, and conversational system code-switching played a major role. \cite{jose2020survey} surveyed a current code-switching dataset and categorized them.

\section{Methodology}
In this section, a detailed description of the dataset used is given along with the detailed steps employed for the pre-processing of the text data and the details of the proposed model architectures.

\subsection{Dataset}
The given dataset comprises of tweets(entries) in code-mixed Hinglish. Each entry contains Twitter handle(s) and links to the corresponding tweet at the beginning and the end of respectively. The entire dataset ( train set + validation set ) has a total of 17000 sentences with an average sentence length of 134.9 characters, has a vocabulary size of 60141. The training set alone has a count of 381970 words making 14594 sentences with 136.2 characters as average sentence length and a vocabulary size of 60115. With a total of 3000 sentences, the validation set had a vocabulary size of 19499 and average sentence length  127.7 characters. The test set also had a total of 3000 sentences vocabulary size of 19331 with an average sentence length of 129.9 characters. \\


\subsection{Pre-Processing}

Removing Noise: The twitter handles and URL links are the main source of noise in the given dataset. Simple regular expressions were created to delete anything which was followed by @ to remove the twitter handles and anything starting with HTTP to remove the URL links.
\vspace{0.2cm}

\noindent Removing Punctuation marks and Special symbols:  All kinds of punctuation marks and special symbols appearing in the dataset were cleared since these characters add no value to text-understanding and in turn induce noise into systems.
\vspace{0.2 cm}

\noindent Removing Stop words: In English, stop words like a, an, the, is, etc. are generally added to make sentences grammatically correct and since they carry minimal value it is apt to remove them so that the focus stays on sentiment determining words. A list of such words is present in the NLTK library. 
\vspace{0.2cm}

\noindent Stemming: Using a snowball stemmer (chosen experimentally) the process of stemming was carried out on the dataset. This was done to bring the words to their respective base forms. This helped a lot in correcting words with unwanted suffixes and skewed spellings which are a common sight in the social media text.   
\vspace{0.2cm}

\noindent Label Encoding: Categorical sentiment values were label encoded as 0,1,2 to negative, neutral, and positive sentiments respectively. This was done to give a numeric representation to the categorical data. 
\vspace{0.2cm}

\noindent Removing high-frequency Hindi words: Unlike English, there does not exist any list of Hindi stop words so such a list was prepared based on Term Frequency (TF) over the entire dataset. The most frequently occurring 1000 words were removed.
\vspace{0.2cm}

\begin{figure}[htp]
    \centering
    \includegraphics[width =16cm]{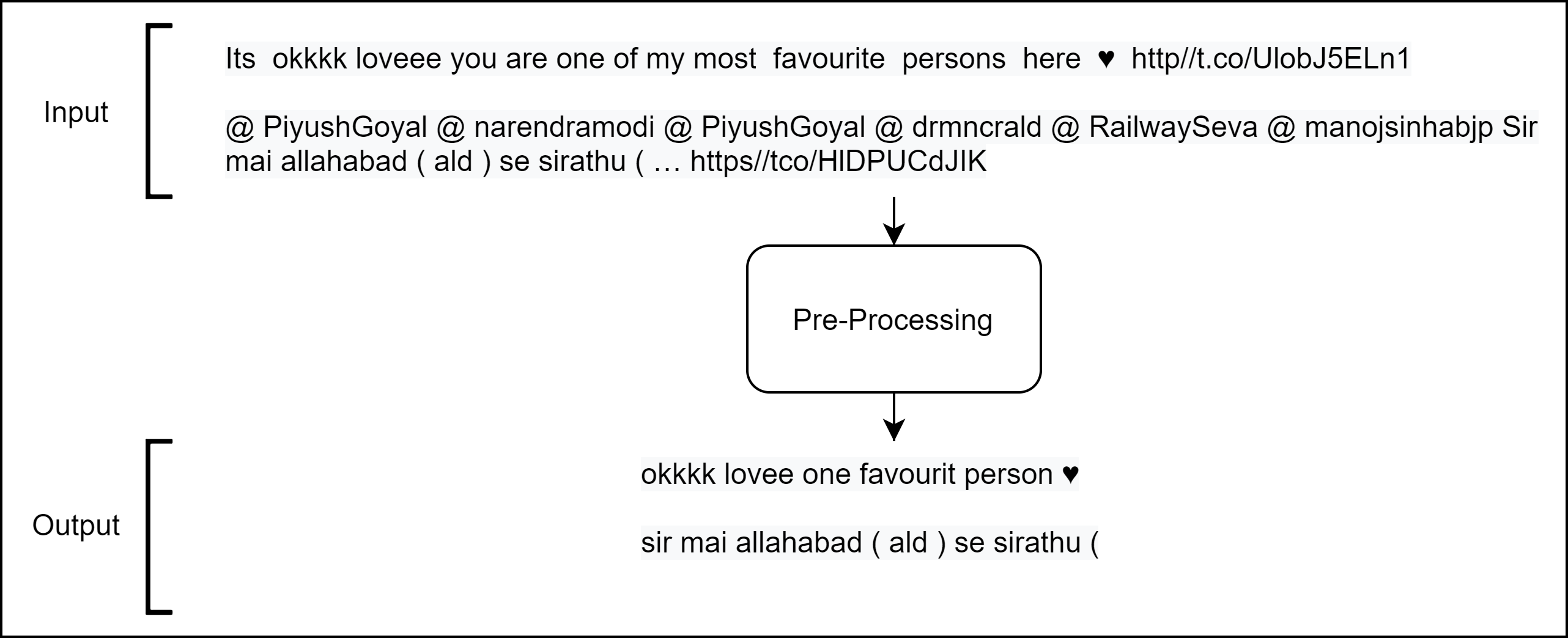}
    \caption{Pre Processing Example}
    \label{fig: (Pre Processing Example)}
\end{figure}

\subsection{Model Architecture}
The proposed system NITS-Hinglish-SentiMix\footnote{
Link to the GitHub repository containing all the codes :- \url{https://github.com/singhnivedita/SemEval2020-Task9}
} is an ensemble model made using on 4 different individual models each of which has been discussed in details in the following section.

\subsubsection{LSTMs}
A simple model which consists of a single layer of {LSTM}\cite{LSTM-1} followed by 2 dense layers, the details are shown in Figure 2(a).

\subsubsection{LSTM + Convolution Layer}

This model contains a convolution layer with kernel size 3 followed by a global max pool layer, LSTM layer and a dense layer\cite{mathur-etal-2018-offend} for which the details are shown in Figure 2(b).

\subsubsection{BiLSTMs}

In this model, a BiLSTM layer followed by a convolution layer with kernel size 3 is employed. The output of this layer is passed through two different layers namely the global average pool and global max pool. The output is concatenated and then passed to a dense layer\footnote{
Link to BilSTM Model :-   \url{https://github.com/tensorflow/docs/blob/master/site/en/tutorials/text/text_classification_rnn.ipynb}
}. Figure 2(c) shows details of Model 3.

\begin{figure}[htp]
    \centering
    \includegraphics[width =16cm,height=20cm]{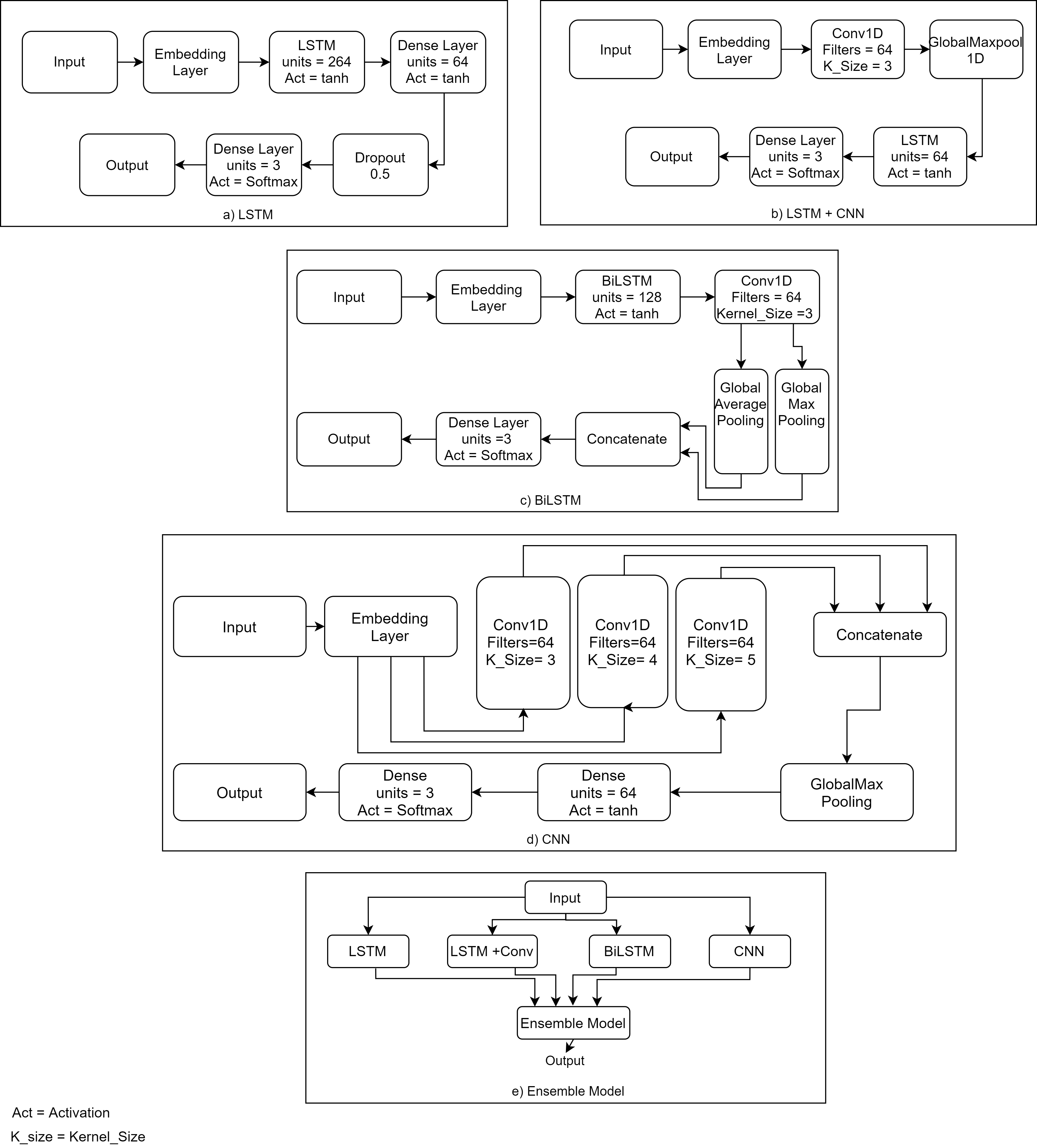}
    \caption{Models}
    \label{fig: (Ensemble Model)}
\end{figure}

\subsubsection{CNN}

This particular model uses 3 different convolution layers with kernel size 3, 4 and 5 connected to the embedding layer. The outputs of each layer is concatenated and then passed to a global max pool layer followed by 2 dense layers \cite{abu-farha-magdy-2019-mazajak} as shown in Figure 2(d).

\subsubsection{NITS-Hinglish-SentiMix Framework}
For better results, an {ensemble model} was constructed to harness the strength of each individual model. On passing the inputs to all models, the outputs were denoted by\[ O_n \text{ where n was the no of the model stated in the previous section}.\]\[O_n = \sum_iO_n^i \text{where i = no of sentences.}\]\[O_n^j \text{ denotes the probability of class j for the }n^{th} \text{ model}\] The final output matrix was calculated using the formula shown below. After calculating, each sentence was assigned a class with the maximum probability.
\[O_final = max(O_{10},O_{20},O_{30},O_{40}), max(O_{11},O_{21},O_{31},O_{41}) ,max(O_{12},O_{22},O_{32},O_{42})\] 
The Figure 2(e) shows the diagrammatic representation of the proposed ensemble model.

\section{Results}
Each of the independent models were trained for 200 epochs with a batch size of 128, vocabulary size of 20000, text sequence length of 50 with sparse categorical loss and learning rate of 0.01. The results have been summarised in the Table no 1.

\begin{table}[ht]
\begin{center}
\begin{tabular}{|l|l|l|l|l|}
\hline \bf Parameter & \bf LSTM &\bf LSTM+Conv &\bf BiLSTM &\bf CNN+Dense\\\hline
Validation F-Score & 0.8413 & 0.8660 & 0.9023  & 0.7757\\
Test F-Score & 0.5640  &  0.5747  & 0.576  & 0.5737\\

\hline
\end{tabular}
\end{center}
\caption{\label{ Individual Model Accuracies-table} Individual Model Accuracies  }
\end{table}

\noindent NITS-Hinglish-SentiMix ensemble model performed optimally with 5 epochs with all the other training parameters remaining same as mentioned above for the respective models. The overall F-Score achieved by NITS-Hinglish-SentiMix on the final submission was 0.617. Table no 2 contains the individual F-score of all the different models on validation and test set.

\begin{table}[ht]
\begin{center}
\begin{tabular}{|l|l|l|l|l|l|}
\hline \bf Parameter & \bf LSTM &\bf LSTM+Conv &\bf BiLSTM &\bf CNN+Dense &\bf NITS-Hinglish-SentiMix \\\hline
Validation F-Score & 0.7720 & 0.8043 & 0.8510  & 0.8360 & 0.7623 \\
Test F-Score & 0.5953  & 0.6077  & 0.5810  & 0.6070  & 0.617  \\
\hline
\end{tabular}
\end{center}
\caption{\label{ NITS-Hinglish-SentiMix validation accuracies for each model-table} NITS-Hinglish-SentiMix validation accuracies for each model  }
\end{table}

\section*{Conclusion}
In this paper, a detailed approach for the sentiment analysis of the code- mixed Hinglish data is described. NITS-Hinglish-SentiMix is an ensemble model over four distinct models that on their own did not fare well with the task which is very much visible in their respective test accuracies on the test data as seen in Table 1. However, it is observed that each individual model was able to catch a particular sentiment exceptionally well hence the decision to adopt an ensemble model. NITS-Hinglish-SentiMix, an ensemble model built on detailed pre-processing and extensive model training, achieves an F-score of 0.617 on the test data. A voted ensemble may be attempted to improve the score further as a future direction of the work.
\bibliographystyle{coling}
\bibliography{semeval2020}

\end{document}